\documentclass[letterpaper, 10 pt, conference]{ieeeconf}  

\IEEEoverridecommandlockouts

\usepackage{amsmath,amsfonts}
\usepackage{algorithmic}
\usepackage{algorithm}
\usepackage{array}
\usepackage[caption=false,font=normalsize,labelfont=sf,textfont=sf]{subfig}
\usepackage{textcomp}
\usepackage{stfloats}
\usepackage{url}
\usepackage{verbatim}
\usepackage{graphicx}
\usepackage{cite}
\usepackage{booktabs,siunitx}
\usepackage{color,soul}
\usepackage{booktabs}
\usepackage{hyperref}
\usepackage{microtype}
 \hypersetup{
     colorlinks=true,
     linkcolor=blue,
     filecolor=blue,
     citecolor = black,      
     urlcolor=cyan,
     }

\title{\LARGE \bf {Tunable Passivity Control for Centralized Multiport Networked Systems}}

\author{Xingyuan~Zhou$^{*}$$^{\dagger}$,~\IEEEmembership{Graduate Student Member,~IEEE,} Peter~Paik$^{*}$,~\IEEEmembership{Graduate Student Member,~IEEE,}\\~and~S.~Farokh~Atashzar,~\IEEEmembership{Senior Member,~IEEE}\vspace{-0.6cm}
\thanks{X. Zhou and P. Paik are with the Dept. of Electrical and Computer Engineering, New York University (NYU), New York, NY, 11201 USA. At the time of the conduction of this research, Atashzar was with NYU. This material is based upon work supported by the US National Science Foundation under grants no \#2121391 and \#2208189. The work is also supported in part by NYUAD CAIR Award \#CG010.The work is also supported in part by GAANN Grant Number P200A210062. The authors would like to acknowledge the support from
MathWorks.}\\
\\
\thanks{$^{\dagger}$ Corresponding author: Xingyuan Zhou ({\tt\footnotesize xz3428@nyu.edu}).}
\thanks{$^{*}$ Peter Paik and Xingyuan Zhou contributed equally to this work.}
}

\begin{document}

\maketitle
\thispagestyle{empty}
\pagestyle{empty}
\bstctlcite{IEEEexample:BSTcontrol}

\begin{abstract}
\sloppy 
Centralized Multiport Networked Dynamic (CMND) systems have emerged as a key architecture with applications in several complex network systems, such as multilateral telerobotics and multi-agent control. These systems consist of a hub node/subsystem connecting with multiple remote nodes/subsystems via a networked architecture. One challenge for this system is stability, which can be affected by non-ideal network artifacts.
Conventional passivity-based approaches can stabilize the system under specialized applications like small-scale networked systems. 
However, those conventional passive stabilizers have several restrictions, such as distributing compensation across subsystems in a decentralized manner, limiting flexibility, and, at the same time, relying on the restrictive assumptions of node passivity. 
This paper synthesizes a centralized optimal passivity-based stabilization framework for CMND systems. It consists of a centralized passivity observer monitoring overall energy flow and an optimal passivity controller that distributes the just-needed dissipation among various nodes, guaranteeing strict passivity and, thus, L2 stability. 
The proposed data-driven model-free approach, i.e., Tunable Centralized Optimal Passivity Control (TCoPC), optimizes total performance based on the prescribed dissipation distribution strategy while ensuring stability. The controller can put high dissipation loads on some sub-networks while relaxing the dissipation on other nodes.
Simulation results demonstrate the proposed frameworks performance in a complex task under different time-varying delay scenarios while relaxing the remote nodes minimum phase and passivity assumption, enhancing the scalability and generalizability.
\end{abstract}
\section{Introduction}
\subsection{Literature Review}
Centralized networked architectures have become vital across domains, comprising centralized hub nodes interconnecting numerous remote nodes through various network topologies. Example applications encompass distributed multi-agent control systems, complex epidemic models, and complex multilateral telerobotic systems. The topic has been extensively investigated in the area of multilateral telerobotics (eg. Single-Leader Multi-Follower (SLMF) architecture) and several control approaches have been designed for this application, including unmanned exploration, haptic communication, multi-agent control, surgical robot training, and more \cite{a1,a2,a3}. In contrast to bilateral dual-node networked systems, which are typically distinguished by one local node (as the sole leader node) and one remote node (as the follower node), centralized multiport networks combine numerous local node(s) and remote node(s). This technology allows several agents-for example, a human or robotic system or a combination-to interact with each other across various locations through network topology. Example architectures for this type of Centralized Multiport Networked Dynamic (CMND) systems can be found in \cite{a1,a2}.
Similar to any networked dynamic architecture, the main challenge is to guarantee and ensure closed-loop stability in the presence of the non-optimum network conditions between each node, besides sensory noises, and actuation faults \cite{i1,i2,i3}. In the last decade, passivity-based controllers have emerged as a solution for both analytical and data-driven control in various applications, including different categories of networked systems \cite{scardovi2010synchronization,4177251,1266772,4287131,zhou2025encodingbiomechanicalenergymargin},. For this, traditionally, wave variable control (WVC)\cite{sun2015wave} and scattering transform control architectures have been proposed and applied in various domains\cite{scardovi2010synchronization}, including bilateral telerobotics. More flexible approaches, such as time-domain passivity control (TDPC) and its variety\cite{JHRyu2004,zhouacc2024,paiktim2025,A_Oliver2023MyoPassivity,B_Oliver2023} and variable structure passivity control (VSPC)\cite{Paik2022}, have been introduced for dual-node systems. In some relatively recent examples \cite{ex4}, an extended TDPC was introduced to the multilateral teleoperation architecture to guarantee stability in the presence of communication time delays. Other examples, such as adaptive controllers based on delay-dependent Lyapunov–Krasovskii functionals, are also proposed \cite{9448498}.
\par
Even though the existing methods can address stability for some of the CMND systems, the fixed decentralized design and the restrictive assumption on the passivity behavior of the nodes limit the scalability and generalizability of such solutions. In other words, the lack of flexibility in coordination between distributed dissipation across the network can lead to inefficient over-dissipation in sub-networks without the ability to allocate control effort for stabilization.

For example, in practical SLMF telerobotics cases, an operator might control a group of robots conducting tasks in parallel (i.e., farming robots, moving or assembly robots). In these cases, when there is a potential challenge to the system's stability, instead of dissipation of the same amount of energy on each robot, there is a need to prescribe allocation of dissipation and control efforts based on the local availability of resources and the importance and significance of local functionality. 
This restricts achievable performance across the network with minimal flexibility to globally adjust or optimize stabilizer performance. The existing fixed structure solutions do not have a way to incorporate node priorities or prescribed weightings into the stabilization strategy while certain critical nodes may require tighter stabilization bounds than others depending on local dynamic uncertainty and pattern of distributed delay. Also, several existing solutions cannot readily compensate for time-varying delays, and offer limited scalability as network complexity increases, and often require models of node dynamics, and most importantly, entail restrictive assumptions of node passivity which may not be guaranteed (for example if a node has a non-minimum phase behavior). \par
\begin{figure}[t]
\vspace{3mm}
\centerline{\includegraphics[width=0.4\textwidth]{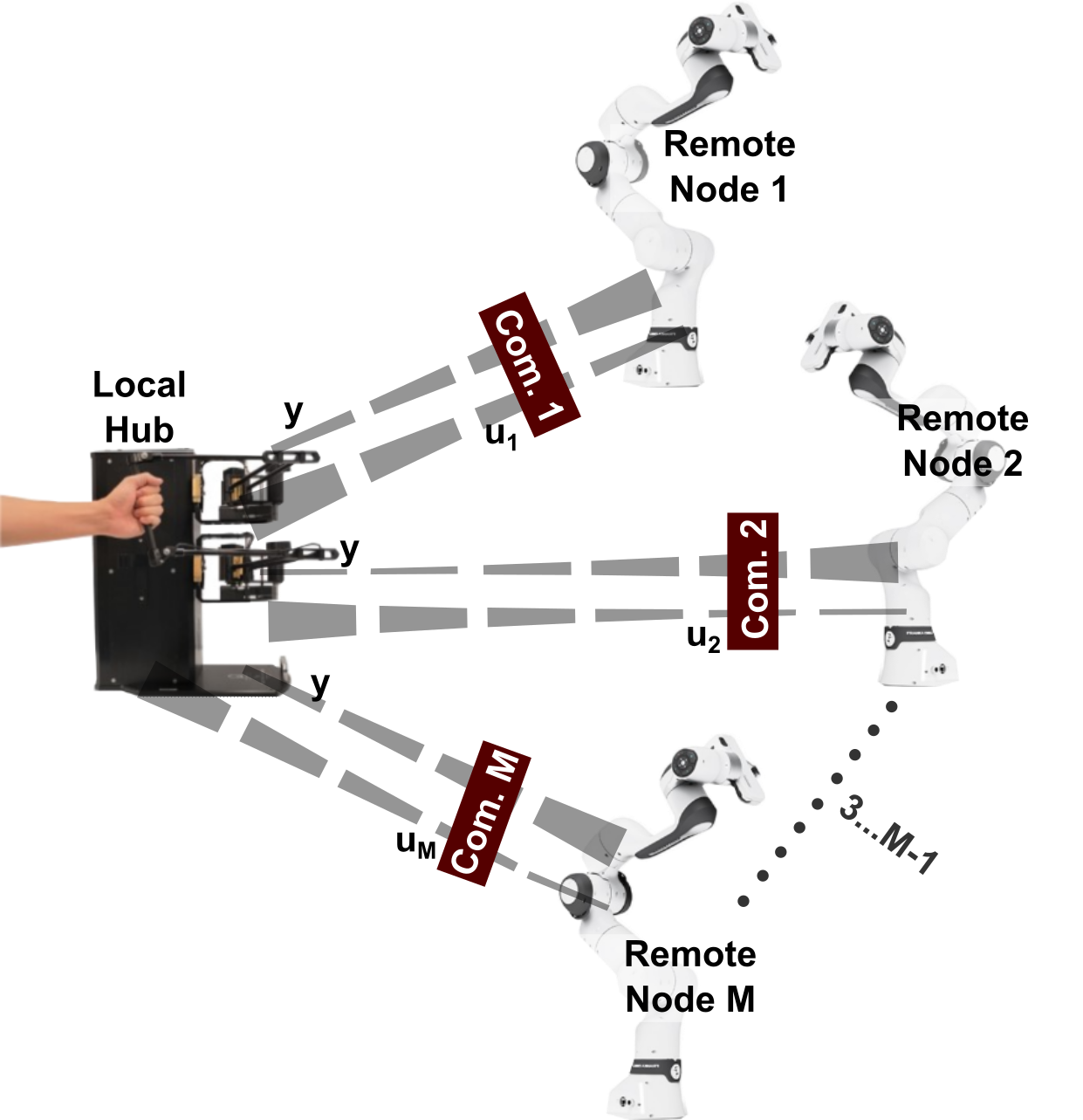}}
\caption{Example network robotic teleoperation architecture of local node communicating with multiple remote nodes.}\vspace{-0.1cm}
\label{example}
\end{figure}

\subsection{The Overview}

These limitations with conventional decentralized stabilization methods motivate the development of a new centralized optimal dissipation framework called Tunable Centralized Optimal Passivity Control (TCoPC). The proposed TCoPC architecture leverages a data-driven model-free approach and overcomes the restrictions of decentralized passivity-based controllers, and enhances performance for generalized centralized networked systems. Specifically, TCoPC incorporates a centralized observer that monitors overall system energy flows throughout the network topology. This provides a holistic view of the energy dynamics to enable ``dissipation planning'' and optimization across nodes in a coordinated way, unlike isolated decentralized methods. The proposed planned dissipation distribution strategically allocates and tunes the stabilization efforts and passivity bounds based on prescribed priorities and weightings given for each node while preserving stability. This allows flexible performance tuning and optimization of the network-wide dissipation. Critical nodes with greater uncertainties or those with more resources can be allocated more dissipation compared to other nodes, ensuring robust stabilization, while excess dissipation is avoided elsewhere to maximize network synchronization. Furthermore, the data-driven model-free nature of TCoPC enhances applicability to complex, uncertain architectures with minimal restrictive assumptions. By directly utilizing real-time data to drive the behavior, stability and performance can be optimized even for unknown, time-varying dynamics and in the presence of time-varying delays, packet losses, node nonpassivity, and non-minimum phase behaviors. In this study, extensive simulation analyses verify that TCoPC significantly improves stability and synchronization for CMND systems, even in the presence of substantial uncertainties, losses, and nonpassive node dynamics, while implementing the prescribed dissipation intensities.  The result indicated that the proposed architecture can induce precise tracking in a series of targetted nodes by reducing the dissipation intensities in those nodes while increasing the dissipation in other subnetworks (customizing dissipation burden on targetted nodes) while guaranteeing the passivity criterion in one shot.

\par

The rest of this paper is organized as follows. In Section II, we provide the preliminaries regarding the definitions of
passivity used in this paper. In Section III, we demonstrate our proposed method. In Sections IV and V, we provide the simulation results
using the proposed architecture. The paper is concluded in Section VII.\vspace{-0.2cm}

\section{Preliminaries}
We present our work for a local node (i.e., hub) communicating with various remote nodes, as can be seen in Fig. \ref{example}. An example of this scenario would be in a network telerobotic architecture where various follower robots are controlled remotely via a single leader robot. The sent signal can be the commanded control velocity to the follower robots, and the returned signals would be the reflective force field from each of the remote follower robots of the teleoperation. The signals are sent and received through individual communication channels, which may be different between each node, non-ideal, and have time-varying delays. To ensure the stability and convergence of the entire interconnection, we use the Passivity Theory approach \cite{def1,def3}. 

\textbf{\textit{Definition 1 (Passivity Condition):}} if there exists a $\beta$ such that the net energy flow of the interconnection is: 

\begin{equation}
    \int_{0}^{T}u(t)^{T}y(t)dt \geq -\beta, ~ ~ \forall{t\ge0},
\end{equation}

then the interconnection is passive. $u(t)$ is the input signal vector to the interconnection (in this case which is the multi-port CMND architecture), and $y(t)$ is the output signal vector of the interconnection. $\beta$ is the initial energy stored in the system, and $\beta=0$ in most applications without the loss of generality. This definition indicates that the net energy flow of the interconnection is positive (i.e., the energy in the system is increasing); thus, the system behaves as a energy-dissipating element and not a energy-generating element. 

The integral is expanded to include the individual energy contributions from each node in the CMND architecture:

\begin{equation}
    \int_{0}^{T}\biggr(u_{hub}(t)y(t)+u_{1}(t)y(t)+...+u_{M}(t)y(t)\biggr)dt \geq 0,
\end{equation}

where $M$ is the total number of remote nodes. This can be written in discrete form 
considering  $t=n\Delta{T}$ where $\Delta{T}$ is the sampling period, as:


\begin{equation} \label{passivity}
    \Delta{T}\biggr(\sum_{n=1}^{N}u_{hub}[n]^{T}y[n]+\sum_{n=1}^{N}\sum_{i=1}^{M}u_{i}[n]^{T}y[n]\biggr) \geq 0, ~~\forall{n\geq1},
\end{equation}

where $n$ is the discrete time index, $N$ is the final timestamp (which can be as $N \rightarrow \infty$), $u_{i}[n]$ is the input vector from the `$i^{th}$' remote node at time stamp $n$, $y[n]$ is the output vector sent to the remote nodes at timestamp $n$, and $u_{hub}[n]$ is the input vector to the local hub at timestamp $n$.  


\textbf{\textit{Definition 2 (Output Strictly Passive System):}} If $u_{hub}[n]$ of the hub is not accessible and cannot be measured during operation (this can be the case for some applications, such as human-telerobot interaction), we utilize the Output Strictly Passive(OSP) condition from strong passivity theory. In this case, we model the behavior of the hub as: 

\begin{equation} \label{osp}
\sum_{n=1}^{N} u_{hub}[n]^{T} y[n] + E(0) \geq \xi \sum_{n=1}^{N} y[n]^{T} y[n]
\end{equation}

where $\xi$ is the inherent Passivity Index (PI) of the local hub\cite{zhou1,zhou2,zhou3,zhouacc2024}.  When this value is positive, the hub acts as an inherent dissipation node (i.e., passive system) in the network that dissipates energy with excess of passivity of $\xi$.


\textbf{\textit{Definition 3: Passivity of CMND Architecture:}} the overall passivity condition of the CMND architecture is obtained by combining \eqref{passivity} and \eqref{osp}:

\begin{equation} \label{final}
    \Delta{T}\biggr(\xi\sum_{n=1}^{N}y[n]^{T}y[n]
    +\sum_{n=1}^{N}\sum_{i=1}^{M}u_{i}[n]^{T}y[n]\biggr) \geq 0, ~~\forall{n\geq1}.
\end{equation}

If the condition in \eqref{final} is met, then the CMND architecture is said to be passive and therefore $L_2$ stable.

\section{Proposed Method}
\begin{figure}[t]
\vspace{4mm}
\centerline{\includegraphics[width=0.42\textwidth]{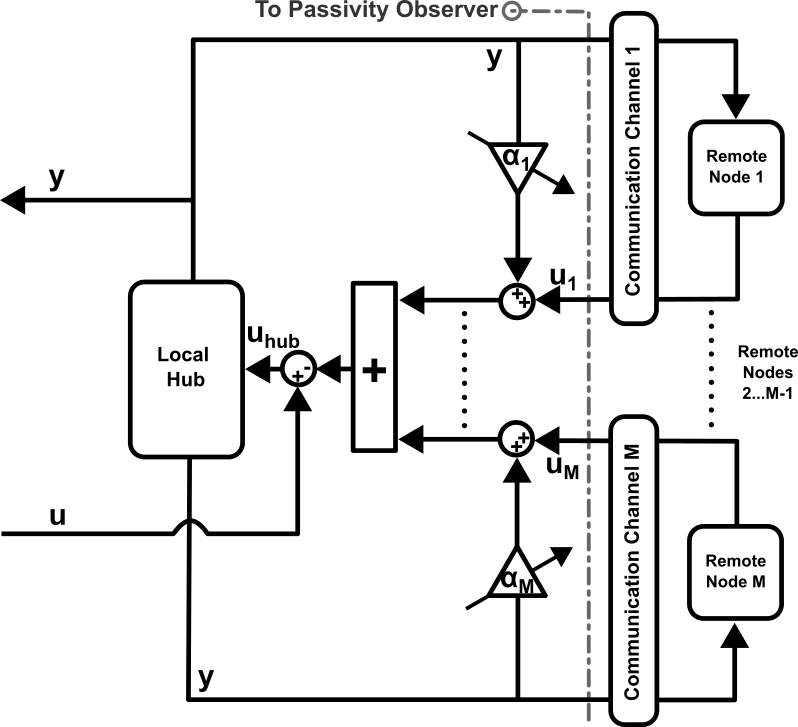}}
\caption{Overall schematic of local hub interconnection with multiple remote nodes in addition to the distributed adaptive dissipative gains of $\alpha_i$.}
\label{arch}
\end{figure}

The proposed architecture is shown in Fig. \ref{arch}. The result is a centralized passivity observer (cPO) and passivity controllers on each nodes (dPC) that incorporates the \textcolor{blue}{PI} of the local node (i.e. the hub) and allows for the tuning/adjusting of the dissipation effort between different nodes, as explained in the rest of this section.

The energy of the entire interconnection at a given discrete integral horizon of $N$ can be written as:

\begin{equation} \label{5}
     E[N] = \Delta{T}\biggr(\xi\sum_{n=1}^{N}y[n]^{T}y[n]+
     \sum_{n=1}^{N}\sum_{i=1}^{M}u_{i}[n]^{T}y[n]\biggr),
\end{equation}

where $\sum_{n=1}^{N}\sum_{i=1}^{M}u_{i}[n]^{T}y[n]$ represents the total energy observed from all remote nodes including their respective communication link, $\xi\sum_{n=1}^{N}y[n]^{T}y[n]$ represents the inherent passive capacity of the hub during this discrete integration window. As mentioned in \eqref{final}, if this energy is positive, then the system is deemed to be passive and stable. {In an extreme scenario, if all the energy injected by the remote nodes is nonpassive, but the passive capacity of the hub is sufficient to compensate for the total nonpassive energy from the remote nodes, the overall system remains passive and, consequently, stable. However, if this balance cannot be maintained, additional dissipation must be introduced to compensate for the passivity deficit and ensure stability. More generally, system passivity can be achieved by incorporating a distributed passivity controller, which is activated upon detecting nonpassivity within the network.} As can be seen in Fig. \ref{arch}, the prescribed dissipation effort is implemented by a series of $\alpha_{i}$ elements in the network. In the presence of the proposed stabilizer, the total energy of the interconnected system will be $ \hat{E}[N]$ for the discrete integral horizon of $N$, and the new passivity condition of the closed-loop system can be written as $ \hat{E}[N] \geq 0$, where we have:  

\begin{equation} \label{passivity2}
  \hat{E}[N] =  E[N]+\Delta{T}\sum_{n=1}^{N}\sum_{i=1}^{M}\alpha_{i}[n]y[n]^{T}y[n].
\end{equation}

In this equation, $\sum_{n=1}^{N}\sum_{i=1}^{M}\alpha_{i}[n]y[n]^{T}y[n]$ is the total energy injected by the distributed passivity controller at node $i$ up to time-step $N$. The $\alpha_{i}[n]$ is the dissipation element provided by the proposed distributed passivity controller for the $i^{th}$ remote node at each timestamp $n$. 

In order to design a causal controller, we calculate the prescribed value of $\alpha_{i}[n]$ at the current timestamp of $n$. For this, the cPO observes all energy sub-elements in \eqref{passivity2} at the current timestamp $n$. As a result, considering \eqref{passivity2}, we have the following relation: \vspace{-0.3cm}


\begin{multline} \label{energy1}
    \hat{E}[n] = E_{obs}[n] +\Delta{T}\sum_{i=1}^{M}\alpha_{i}[n]y[n]^{T}y[n]\\
 \text{and    } E_{obs}[n]=  E[n] + \Delta{T}\sum_{k=1}^{n-1}\sum_{i=1}^{M}\alpha_{i}[k]y[k]^{T}y[k],
\end{multline}

where $\Delta{T}\sum_{k=1}^{n-1}\sum_{i=1}^{M}\alpha_{i}[k]y[k]^{T}y[k]$ is the total energy that has been injected by the distributed passivity controller at timestamp $n$. As a result,  $E_{obs}[n]$ is the observable energy at timestamp $n$.



Based on access to the observable energy $E_{obs}[n]$, we can design the causal adaptive vector of stabilizers $\alpha_{i}$ for all nodes at time $n$ in a way that can guarantee \eqref{passivity2}  is positive semidefinite, based on definition 1, the entire interconnected system in the presence of the proposed distributed stabilizer is passive. 

For this, we reorganize the achieved passivity condition so that we can isolate the to-be-designed stabilization component at time $n$ for each node. From \eqref{passivity2}, and \eqref{energy1}, we have the system is stable if: 

\begin{equation} \label{energy2}
    E_{obs}[n]+\Delta{T}\sum_{i=1}^{M}\alpha_{i}[n]y[n]^{T}y[n]\geq0.
\end{equation}

Based on the relation between $E_{obs}[n]$ $\alpha_{i}[n]$, the total passivity condition \eqref{energy2} can be rewritten to the following form:

\begin{equation} \label{9}
    \text{The system is passive if } A^{T}S \geq \frac{-E_{obs}[n]}{\Delta{T}}
\end{equation}

In \eqref{9}, we have $A=[\alpha_{1},\alpha_{2},...,\alpha_{M}]^T$ is the vector containing the adaptive dissipation elements for each remote node, and $S$ is the vector containing the value $y[n]^{T}y[n]$ for each remote node. Thus, by taking the vector product $A^{T}S$, we obtain $\sum_{i=1}^{M}\alpha_{i}[n]y[n]^{T}y[n]$. It is an underdetermined system with infinite solutions. In order to be able to tune the dissipation contribution for each nodes as prescribed by the designer and to guarantee the passivity condition, we formulate the passivity condition as an optimization problem:

\begin{equation}
   \label{10v} \min_{A}J(A)=\frac{1}{2}\min_{A}A^{T}QA ~~ \text{subject to} ~ A^{T}S=\frac{-E_{obs}[n]}{\Delta{T}}
\end{equation}

In this way, we obtain a solution that solves for the adaptive dissipation vector that provides minimal dissipation while guaranteeing the passivity by being subject to the constraint (\ref{9}). 
In \eqref{10v}, the weighting matrix `$Q$' is a symmetric positive semi-definite matrix representing the associated penalty for modifying different $\alpha_{i}$. When the associated diagonal element of $Q$ is large, there is a larger penalty for modifying the corresponding $\alpha$, thus effectively reducing the dissipation applied for that  node. However, when the associated diagonal element is small, there is a smaller penalty, thus increasing the dissipation applied. The optimization problem can be solved using the Lagrange Multiplier Method \cite{lagrange} as:

\begin{equation}
    \mathcal{L}(A,\lambda) = \frac{1}{2}A^{T}QA + \lambda\biggr(A^{T}S+\frac{E_{obs}[n]}{\Delta{T}}\biggr)
\end{equation}

To find the optimum, we can compute the gradients as:

\begin{equation}
    \frac{\delta\mathcal{L}}{\delta{A}}=0 ~ \text{and} ~ \frac{\delta\mathcal{L}}{\delta\lambda}=0.
\end{equation}

Solving for the gradients yield the following:
\begin{equation}
    \frac{\delta\mathcal{L}}{\delta{A}}=QA+S\lambda=0
\end{equation}
\begin{equation}
    \frac{\delta\mathcal{L}}{\delta{\lambda}}=A^{T}S+\frac{E_{obs}[n]}{\Delta{T}}=0
\end{equation}

From the first gradient ($\frac{\delta\mathcal{L}}{\delta{A}}=0$), we obtain the relationship:
\begin{equation} \label{grad1}
    A=-Q^{-1}S\lambda
\end{equation}

And by substituting \eqref{grad1} into the second gradient ($\frac{\delta\mathcal{L}}{\delta\lambda}=0$), we obtain the relationship:
\begin{equation} \label{grad2}
    \lambda = (S^{T}Q^{-1}S)^{-1}\frac{E_{obs}[n]}{\Delta{T}}
\end{equation}

And by combining the solution from \eqref{grad2} and \eqref{grad1}, we obtain the  solution to the optimization problem as given in the following:

\begin{equation}
\label{C1}
    A=-Q^{-1}S(S^{T}Q^{-1}S)^{-1}\frac{E_{obs}[n]}{\Delta{T}}=-S^{\dagger}\frac{E[n]}{\Delta{T}}
\end{equation}

Thus, the design of the stabilizer is defined by \eqref{C1}. The term $S^{\dagger}=Q^{-1}S(S^{T}Q^{-1}S)^{-1}$ can be viewed as a weighted pseudoinverse of $S$. Thus, we obtain the necessary dissipation coefficients $\alpha_{i}$ that need to be applied to each   node to optimally stabilize the system, while the $Q$ matrix allows for tuning of the dissipation effort as prescribed by the designer. 

To summarize the proposed optimal passivity design, it can be said that the proposed stabilizer is comprised of two components: a centralized passivity observer that observes the energy of the entire interconnection and a distributed passivity controller that adaptively and optimally injects distributed dissipation to the node's signals while following a preferred dissipation priority prescribed by the designer through matrix $Q$. The centralized PO observes the system energy represented by \eqref{energy}. The distributed PC applies the necessary dissipation at the most recent time-step $n$ to each node accordingly. The design is summarized as follows: 

\begin{multline}   A[n] = \begin{cases}
    0, & \text{$E_{obs}[n] \geq 0$}.\\\\
    -S^ {\dagger}\frac{E_{obs}[n]}{\Delta{T}}, \text{where: 
   } \\ S^{\dagger} = Q^{-1}S(S^{T}Q^{-1}S)^{-1}  & \text{otherwise.}
  \end{cases}
\end{multline}

In the next two sections, the performance of the proposed stabilizer (i.e., TCoPC) is evaluated for a system with three remote nodes.

\section{Simulation Setup}
To evaluate and verify the proposed TCoPC framework, we implement the stabilizer in a multilateral teleoperation architecture with a single local hub communicating with three different remote nodes. For telerobotic architectures, the signals sent to the remote nodes are the commanded velocities `$v$', and the received signals are the force feedback `$f$' resulting from the interaction of the follower robots with a remote dynamic environment (see our previous publications for more details on 2-port bilateral telerobotic systems \cite{OSP1,Paik2022}). The simulated parameters of the nodes'  dynamics and the variable communication delays are shown in Table \ref{table}. 

\begin{table}
\vspace{3mm}
\centering
  \caption{Simulation Parameters}\label{table}
  \begin{tabular}{ |c|c| }
  \hline
  \textbf{Parameter} & \textbf{Value} \\ 
  \hline\hline
  &\\[-0.8em]
  Local Hub & $Z_{local}$ = $\frac{s}{0.5s^{2}+15s+1}$
  \\[0.5em] 
  \hline
  &\\[-0.8em]
  Remote Node 1 & $Z_1$ = $10s+5+400/s$\\[0.5em] 
  \hline
  &\\[-0.8em]
  Remote Node 2 & $Z_2$ = $-10s-5-400/s$\\[0.5em] 
  \hline
  &\\[-0.8em]
  Remote Node 3 & $Z_3$ = $-20s-10-800/s$\\ [0.5em] 
  \hline
  &\\[-0.8em]
  Round-trip Delay 1 & $t_{d1}$ = $0.05(0.25\sin{(20t)})+0.05 (second)$\\ [0.5em] 
  \hline
  &\\[-0.8em]
  Round-trip Delay 2 & $t_{d2}$ = $0.1(0.25\sin{(20t)})+0.1 (second)$\\[0.5em] 
  \hline
  &\\[-0.8em]
  Round-trip Delay 3 & $t_{d3}$ = $0.15(0.25\sin{(20t)})+0.15 (second)$\\[0.5em] 
  \hline
  \end{tabular}
\end{table}

It should be noted that we utilize asymmetric varying time delays between the remote nodes and the hub. Classically, this would impose three layers of complexity: (a) unknown delay, (b) time-variable delay, (c) different delays in the network links causing asymmetry and asynchrony. However, the design of the proposed stabilizer does not make any assumptions on these items. In other words, the stabilizer is delay-independent and able to function properly when the communication links between the remote nodes and hub present different characteristics. 

Moreover, the proposed stabilizer does not make any assumption on the passivity behavior of the remote nodes, which relax a common assumption that exist in some existing solutions (such as wave variable control that would need the nodes to be passive). To test this functionality, the second and the third nodes are designed to be nonpassive, and the first node is designed to be passive to demonstrate the generality of the proposed stabilizer.


The simulation is conducted with and without the proposed stabilizer in the loop to assess the effectiveness of TCoPC architecture. For the stability test, we provide an impulse signal as the input and measure the energy and power from each remote node and also the state trajectories over time for each node. It should be noted that since a telerobotic example is considered here, the states would be position and velocity; also, the controller would inject damping (to dissipate), which means that the controller would modify the force in the feedback channel.

To test the functionality of the proposed TCoPC approach in distributing the dissipation load (in a prescribed manner) between different nodes, we test the stabilizer for different values of the matrix `$Q$.' As a result, three cases for the matrix $Q$ are simulated that would provide three different distribution patterns of dissipation, as explained in the next section. For this experiment, to better visualize the behavior, instead of the impulse response, we applied an input signal of: $u(t)=20\big(\sin{(\pi{t})}+\sin{(0.5\pi{t})}\big)$. In the first case, the energy dissipation load is spread equally between all nodes, which is achieved by a $Q$ equal to the identity matrix. In the second case, we apply all the dissipation effort onto the second node's  by making the second diagonal element in the $Q$-matrix very small ($Q=\text{Diag}(1,0.0001,1)$). In the third case, we apply minimal dissipation effort onto the second node's signal by making the second diagonal element very large ($Q=\text{Diag}(1,10000,1)$). To evaluate the functionality of the controller, we plotted the signals before and after the dissipation induced by the distributed PC. We also plot the associated $\alpha$-values for each node and the energy dissipation load implemented by each node's passivity controller.

\section{Results}
We present the results of the proposed stabilization framework and some example case studies resulting from different choice of $Q$-matrices to evaluate the performance of the proposed framework. The analysis compares the system impulse response with and without the proposed stabilizing architecture. The case studies compare the results from tuning the $Q$-matrix such that we have (1) an equal distribution of energy dissipation among three nodes, (2) all energy dissipation effort loaded on the second node, and (3) minimal energy dissipation effort loaded on the second node. This would show the functionality of the proposed framework and the tunable dissipation distribution strategy.

\begin{figure}[t]
\centerline{\includegraphics[width=0.5\textwidth]{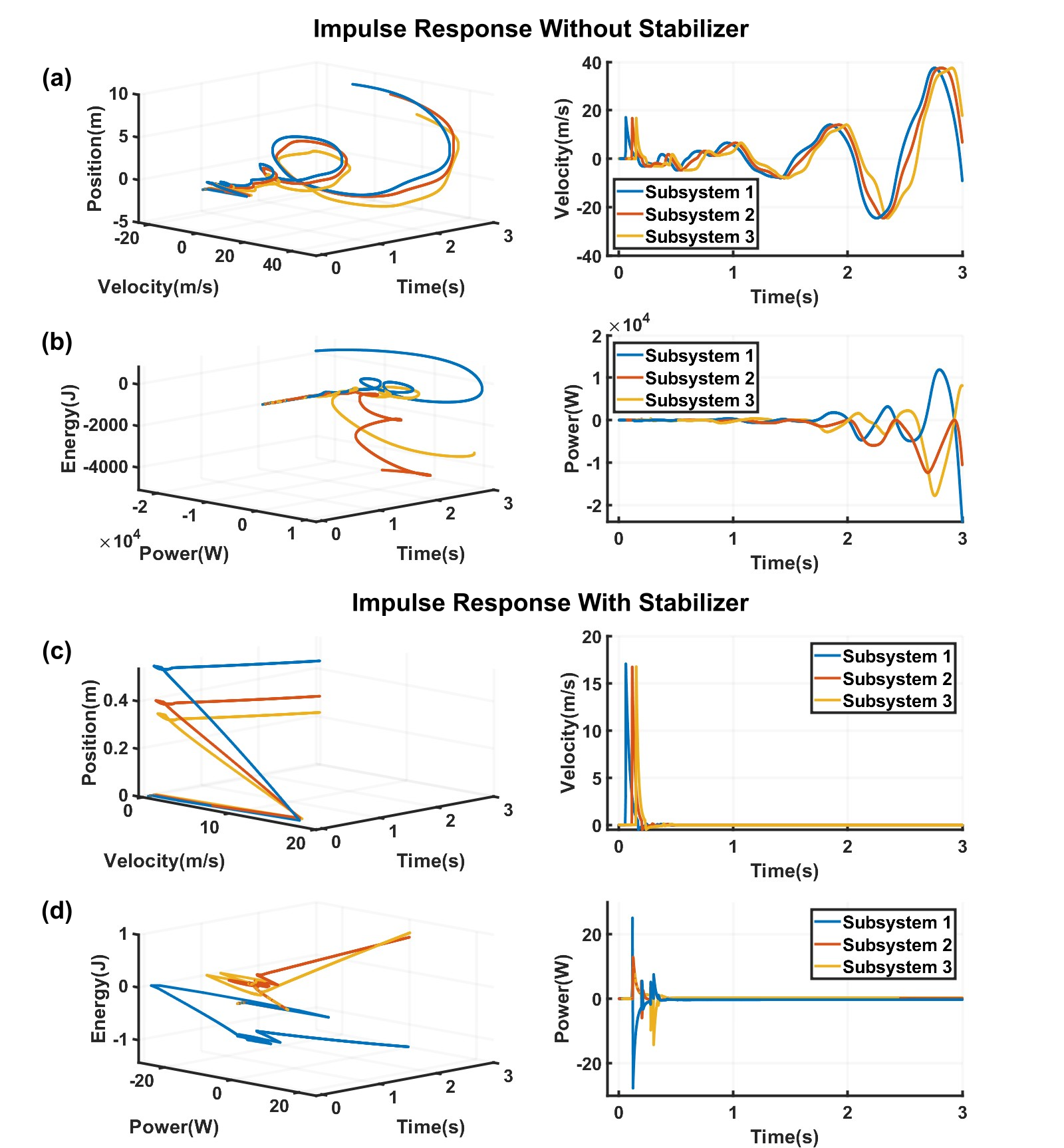}}
\caption{Impulse response for the system with (bottom) and without (top) the proposed stabilizer. (a) states' (velocity and position) trajectories over time for each node without a stabilizer. (b) Energy vs. Power over Time for each node without stabilizer. (c) states' (velocity and position) trajectories over time for each node in the presence of the stabilizer. (d) Energy vs. Power vs. Time plot for each node in the presence of the stabilizer.}
\label{stability}
\end{figure}

\subsection{Stability Validation}
The impulse response of the system with and without the stabilizer is presented in Fig. \ref{stability}. The velocity and position profiles, as well as the power and energy level of the system, are analyzed. In the absence of the controller, since the system is subject to variable communication delays and nonpassive remote node dynamics, the passivity of the interconnected system is challenged. Thus, as can be seen in Fig. \ref{stability}(a), a diverging pattern can be observed for the velocity and position of all nodes over time, highlighting an unstable response of the interconnected system given an impulse input.   Likewise, Fig. \ref{stability}(b) shows the energy-power plot over time for the system without the stabilizer. The power and energy also diverge and become increasingly negative, showing that the passivity condition is violated and the system is unstable. On the other hand, the impulse response for the system in the presence of the proposed stabilizer is shown in Fig. \ref{stability}(c) and \ref{stability}(d). Unlike the previous case, the position and velocity converge over time, showing that the system reaches a steady-state position and zero velocity and zero power given an impulse input, and the energy level of the system also reaches steady-state. The results show that the proposed stabilizer is able to impose the passivity condition and thus stabilize the interconnected system.

\subsection{Case Study: Adjustable Dissipation Distribution}
In the case study, we evaluate the stabilizer's behavior when assigning three different `$Q$'-matrices: 
\subsubsection{Case 1}
 In this case, $Q$ is identity matrix, which allows for equal distribution of energy dissipation among all nodes;
 \subsubsection{Case 2} 
 In the second case, the second diagonal element of the $Q$ matrix, is reduced to a small value, which mathematically would correspond to putting all the energy dissipation effort on the second node; 
 \subsubsection{Case 3}
In the third case, the second diagonal element of the $Q$-matrix is increased to large value, which relatively places minimal energy dissipation effort on the second node. This behavior is expected because the Q-values represent the penalty associated with having a large $\alpha$; by reducing the penalty, the PC focuses on loading the dissipation effort to the other two nodes using their associated $\alpha$. \par
\begin{figure}[t]
\centerline{\includegraphics[width=0.5\textwidth]{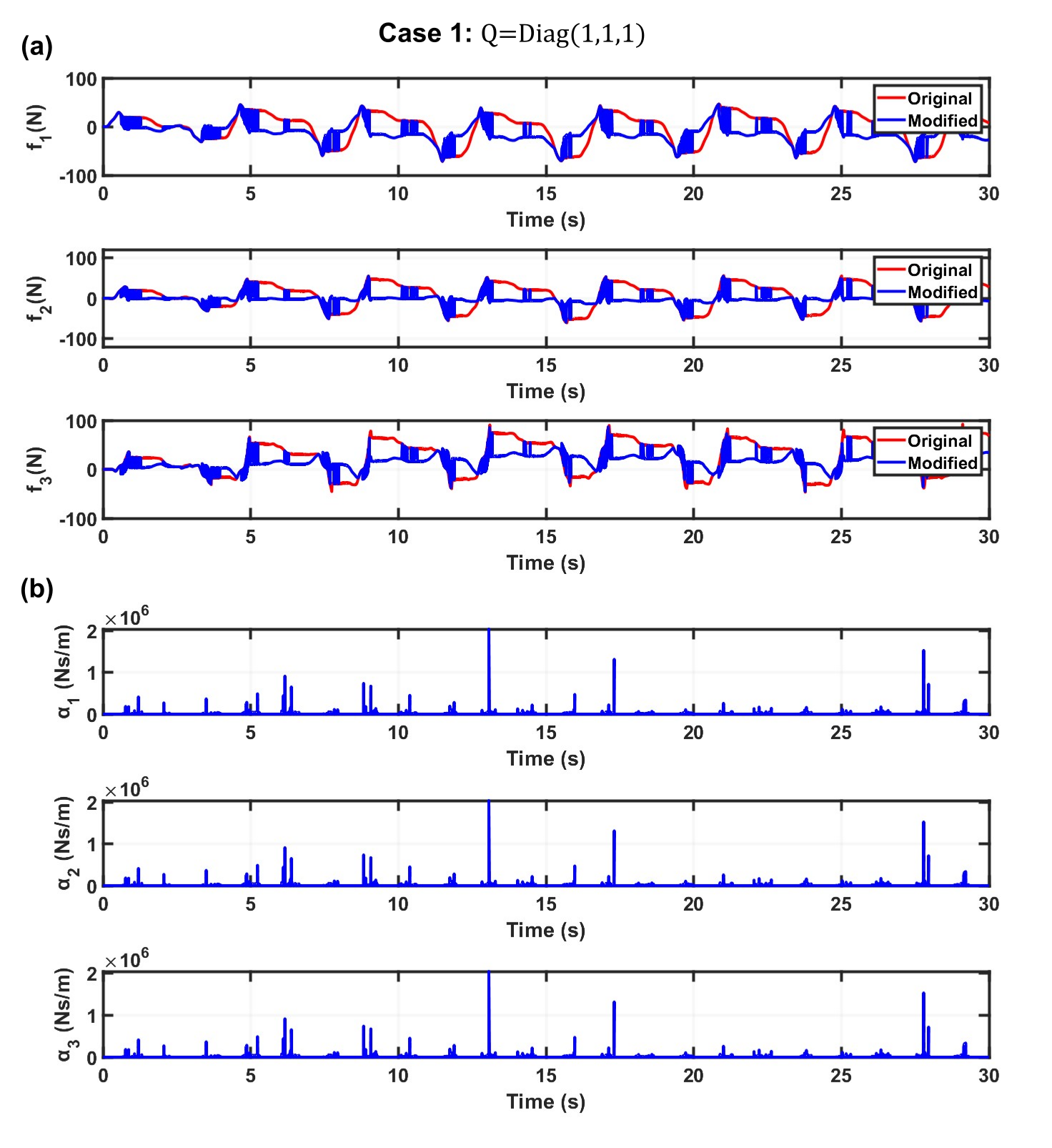}
\vspace{-0.5cm}}
\caption{Case 1 ($Q=\text{Diag}(1,1,1)$): (a) Force modification (before and after the controller)  for each node; (b) the $\alpha$ plots (showing the activations of the controller).}
\label{case1}
\vspace{-0.5cm}
\end{figure}
\begin{figure}[t]
\centerline{\includegraphics[width=0.5\textwidth]{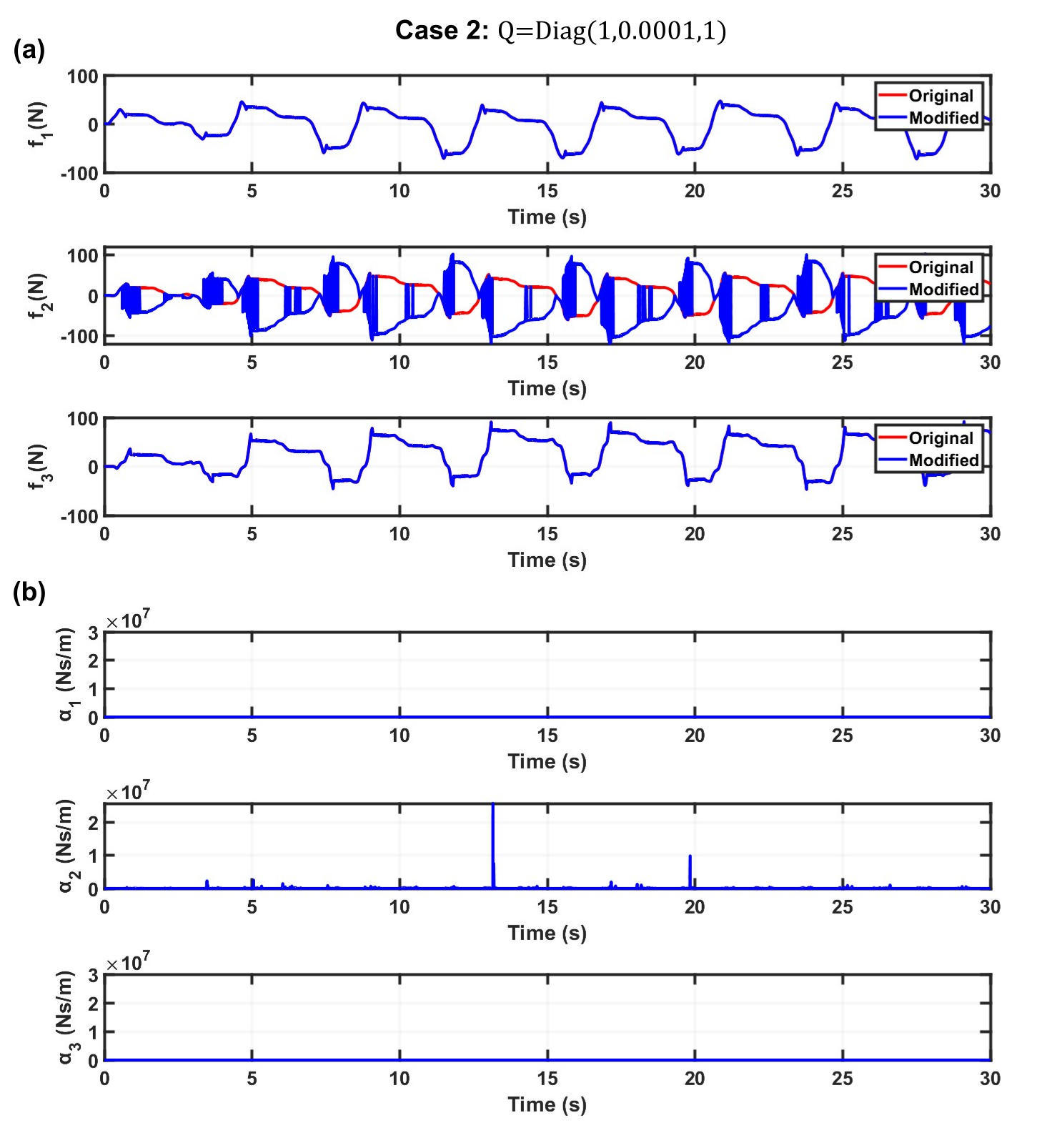}}
\vspace{-0.5cm}
\caption{Case 2 ($Q=\text{Diag}(1,0.0001,1)$): (a) Force modification (before and after the controller)  for each node; (b) the $\alpha$ plots (showing the the activations of the controller).}
\label{case2}
\vspace{-0.2cm}
\end{figure}
\begin{figure}[t]
\centerline{\includegraphics[width=0.5\textwidth]{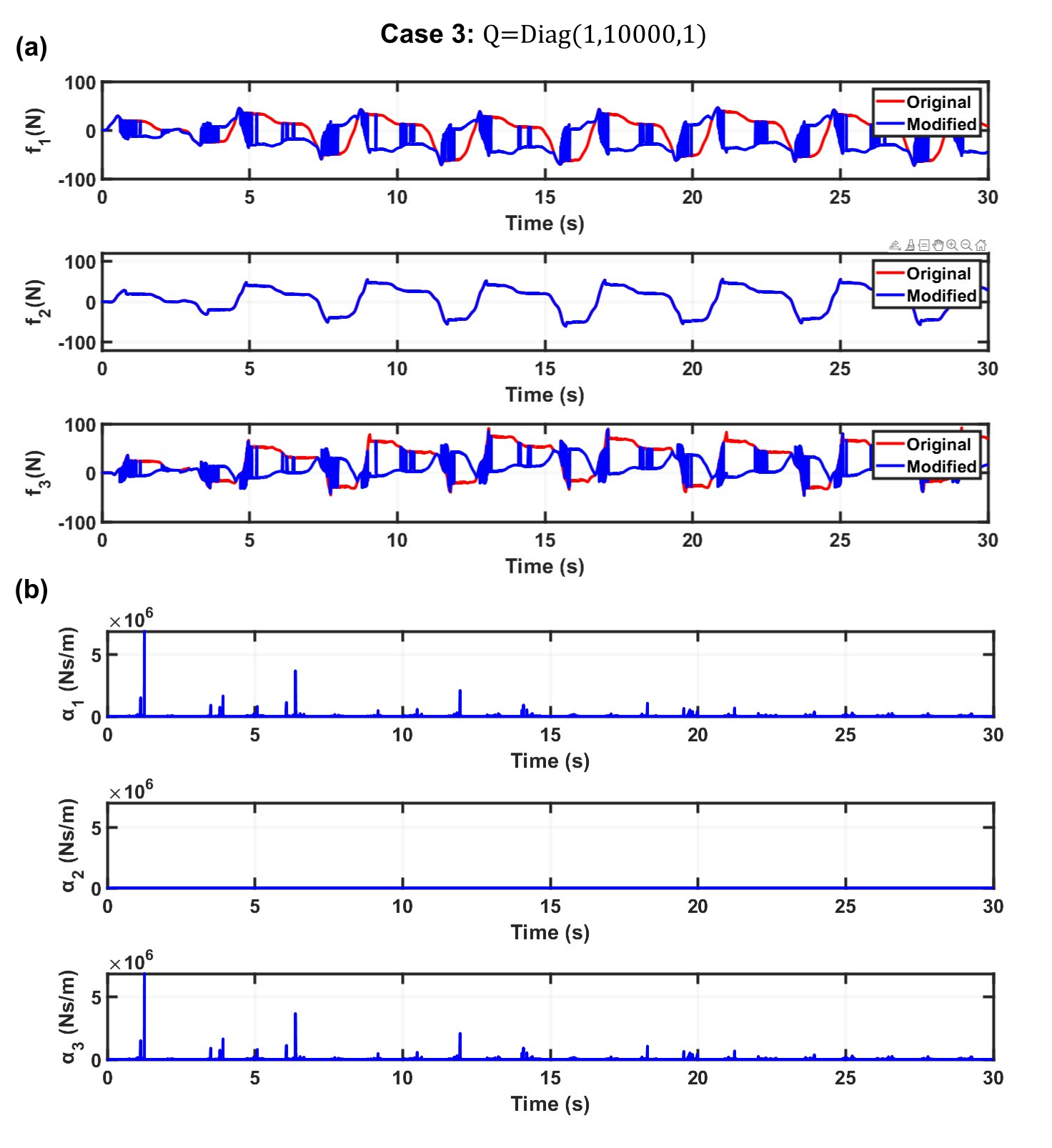}}
\vspace{-0.5cm}
\caption{Case 3 ($Q=\text{Diag}(1,10000,1)$): (a) Force modification (before and after the controller)  for each node; (b) the $\alpha$ plots (showing the activations of the controller).}
\label{case3}
\vspace{-0.5cm}
\end{figure}

It should be noted that since the case study is designed for teleoperation systems, the controllers would modify the force coming back; thus, in order to show the behavior of the controller and the intensity of dissipation, the force profile before and after the controller is compared. A high-intensity dissipation would be translated into relatively higher changes of force signal. In the results, we also directly plotted the $\alpha$, which shows the activation of the controller. The case study results are shown in Figures \ref{case1}, \ref{case2}, and \ref{case3}. 

\begin{figure}[t]
\centerline{\includegraphics[width=0.5\textwidth]{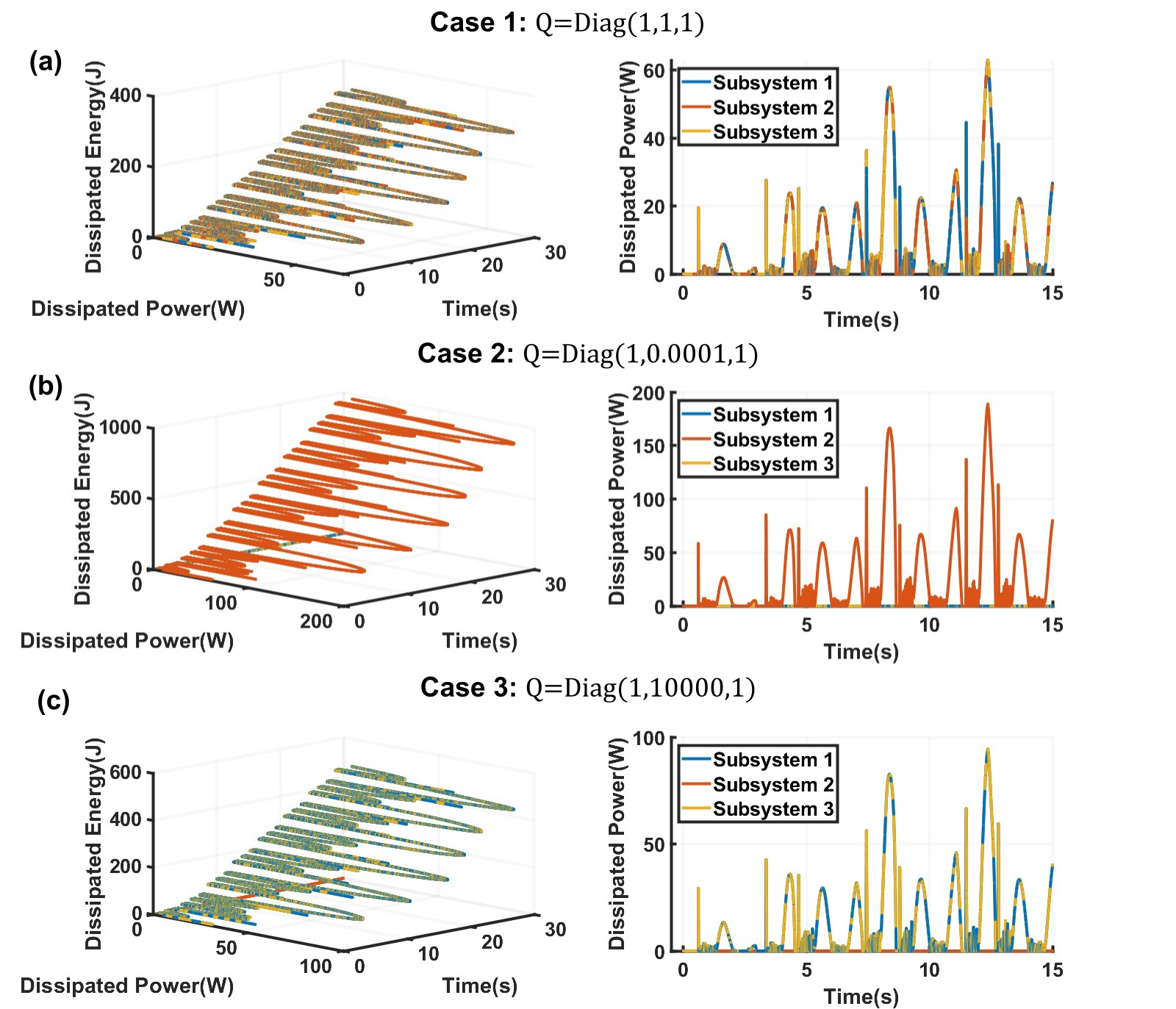}}
\caption{Dissipated Energy enacted by the stabilizer for each node in Case 1 (a), Case 2 (b), and Case 3 (c)}
\label{energy}
\vspace{-0.5cm}
\end{figure}

As can be seen in Fig. \ref{case1}(a) for Case 1, there is a dissipation effort applied on all three nodes, which modifies the original signal value. Fig. \ref{case1}(b) shows the associated $\alpha$ activation on each node. As can be seen, the applied dissipation is comparable for all three nodes. This is due to the design of the $Q$-matrix, which is an identity matrix in this example. 

As can be seen in Fig. \ref{case2}(a) for Case 2, the original force is minimally modified by the controller for node 1 and 3. This is because in Case 2, the $Q$-matrix is designed so that the dissipation effort is placed mostly on the second node. Thus, as can be seen in this figure,  the force is heavily modified for the second node. This is reflected in the $\alpha$ plots in Fig. \ref{case2}(b). The magnitude of the $\alpha$ for the second node is much higher than the magnitude of the $\alpha$ for the first and third nodes. 

As mentioned before, in Case 3 , the $Q$-matrix is designed to have minimal dissipation effort on the second node by increasing the penalty associated with the $\alpha$ of node 2 to large value. As a result, as seen in Fig.\ref{case3}(a), the force on the second node is minimally modified, while the forces on the first and third nodes are heavily modified. These results are reflected in the plots of each node's $\alpha$ Fig. \ref{case3}(b).

We also show the prescribed energy dissipation effort for each node in Fig. \ref{energy}. As can be seen, for Case 1, the prescribed energy dissipation effort for all three nodes is equivalent. In Case 2, the energy dissipation effort is highest for node 2, while nodes 1 and 3 have negligible prescribed energy dissipation. In Case 3, the energy dissipation is minimal for the second node and is similarly high for the first and third nodes. 

Therefore, the results above showed that by modifying the diagonal elements of the $Q$-matrix in the proposed adjustable optimal passivity controller; the designer can intuitively tune the prescribed energy dissipation effort applied on each node with no concern on the overall passivity (thus stability) of the interconnected system. The results in Figures \ref{case1}, \ref{case2}, and \ref{case3} support the expected behavior of the synthesized stabilizer.

\section{Conclusion}
In this paper, we propose a data-driven model-free approach named Tunable  Centralized Passivity Control (TCoPC), for Centralized Multiport Networked Dynamic (CMND) systems. The proposed approach 
is composed of a centralized passivity observer and an optimal passivity controller that allocates the just-needed dissipation among various sub-networks based on the prescribed dissipation distribution strategy to optimize the overall system's performance while guaranteeing the system's $L_2$ stability. In this paper, the mathematical derivation of the proposed approach is provided, as well as a systematic simulation. The simulation result demonstrated that the proposed approach allows the user to intuitively tune the dissipation effort while achieving high performance for desired nodes and maintaining system stability. In addition, the method can perform optimally while relaxing conservative assumptions, such as including time-varying delay between nodes and allowing remote node's non-minimum phase and nonpassive properties. Due to these admirable contributions, the proposed approach has great application in multi-agent control, multilateral telerobotics, and complex epidemic networks.

\section{Acknowledgements}
The authors would like to acknowledge the help of Mr. Sarmad Mehrdad and Mr. Ahmadreza Ahmadjou.

\bibliographystyle{IEEEtran}
\bibliography{ref}

\end{document}